\documentclass[sigconf]{acmart}
\usepackage{bbding}
\usepackage{multirow} 
\usepackage{balance} 
\AtBeginDocument{%
  }

\copyrightyear{2026}
\acmYear{2026}
\setcopyright{cc}
\setcctype{by}
\acmConference[MM '26] {Proceedings of the 34th ACM International Conference on Multimedia}{November 10--14, 2026}{Rio de Janeiro, Brazil}

\acmBooktitle{Proceedings of the 34th ACM International Conference on Multimedia (MM '26), November 10--14, 2026, Rio de Janeiro, Brazil}

\acmISBN{979-8-4007-2213-4/2026/11}
\acmDOI{10.1145/3767308.3835030}

\makeatletter
\renewcommand{\@fnsymbol}[1]{%
  \ifcase#1
  \or \textdagger
  \or \textasteriskcentered
  \or \textdaggerdbl
  \or \S
  \or \P
  \or \|
  \else\@ctrerr
  \fi
}
\makeatother
\begin{document}

\title{ProtoHGF-Net: Prototype HyperGraph Fusion with Intra-modal Calibration for RGBT Object Detection}
\author{Xiangqi Chen}
\authornote{Xiangqi Chen and Zhonglong Zheng are with the
Zhejiang Key Laboratory of Intelligent Education Technology
and Application, Zhejiang Normal University.}
\orcid{0000-0003-3678-6685}
\affiliation{%
  \institution{Zhejiang Normal University}
  \city{Jinhua}
  \state{Zhejiang}
  \country{China}
}
\email{zjnu\_cxq@zjnu.edu.cn}

\author{Xiuling Zhang}
\orcid{0000-0002-7203-2007}
\affiliation{%
  \institution{National University of Defense Technology}
  \city{Changsha}
  \state{Hunan}
  \country{China}
}
\email{xiuling@nudt.edu.cn}

\author{Chengzhuan Yang}
\orcid{0000-0002-6675-9074}
\affiliation{%
  \institution{Zhejiang Normal University}
  \city{Jinhua}
  \state{Zhejiang}
  \country{China}
}
\email{czyang@zjnu.edu.cn}

\author{Li Zhao}
\orcid{0000-0001-5787-2705}
\affiliation{%
  \institution{Zhejiang Normal University}
  \city{Jinhua}
  \state{Zhejiang}
  \country{China}
}
\email{zhaoli2023@zjnu.edu.cn}

\author{Dawei Zhang}
\orcid{0000-0002-7593-1593}
\affiliation{%
  \institution{Zhejiang Normal University}
  \city{Jinhua}
  \state{Zhejiang}
  \country{China}
}
\email{davidzhang@zjnu.edu.cn}

\author{Yanchao Wang}
\orcid{0009-0007-9493-5411}
\affiliation{%
  \institution{Zhejiang Normal University}
  \city{Jinhua}
  \state{Zhejiang}
  \country{China}
}
\email{yanchaowang@zjnu.edu.cn}

\author{Liyuan Chen}
\orcid{0000-0001-7816-5874}
\authornote{Co-corresponding authors.}
\affiliation{%
  \institution{National University of Defense Technology}
  \city{Changsha}
  \state{Hunan}
  \country{China}
}
\email{chenliyuan0905@nudt.edu.cn}

\author{Hua Wang}
\orcid{0000-0002-8465-0996}
\affiliation{%
  \institution{Victoria University}
  \city{Melbourne}
  \state{Victoria}
  \country{Australia}
}
\email{hua.wang@vu.edu.au}

\author{Hao Peng}
\orcid{0000-0003-0586-7132}
\affiliation{%
  \institution{Zhejiang Normal University}
  \city{Jinhua}
  \state{Zhejiang}
  \country{China}
}
\email{hpeng@zjnu.edu.cn}

\author{Zhonglong Zheng}
\authornotemark[2]
\authornotemark[1]
\orcid{0000-0002-5271-9215}
\authornote{Zhonglong Zheng is also with the
China--Mozambique Belt and Road Joint Laboratory
on Smart Agriculture, Zhejiang Normal University.}
\affiliation{%
  \institution{Zhejiang Normal University}
  \city{Jinhua}
  \state{Zhejiang}
  \country{China}
}
\email{zhonglong@zjnu.edu.cn}
\renewcommand{\shortauthors}{Xiangqi Chen et al.}

\begin{abstract}
 RGB-Thermal (RGBT) object detection enables robust perception in complex scenes by leveraging the complementary strengths of visible textures and thermal cues. However, existing methods mainly rely on dense cross-modal interactions over full-resolution features, which inevitably introduce background interference and hinder the learning of target-relevant representations. In this paper, we propose the Prototype HyperGraph Fusion Network (ProtoHGF-Net), a novel framework that redefines cross-modal fusion as prototype-level semantic interaction rather than the dense cross-modal interaction paradigm. Specifically, we design Prototype HyperGraph Fusion to perform cross-modal interaction in a compact prototype-level semantic space. 
This design enables more selective fusion among target-relevant prototypes. To support this prototype-level fusion, we propose Teacher-Mask Calibration Distillation, which calibrates modality features before fusion using modality-specific teachers and target-aware masks. This strategy suppresses backgrou-
nd-dominant responses and produces more target-focused features. Extensive experiments on DroneVehicle, DVTOD, and FLIR demonstrate that ProtoHGF-Net achieves state-of-the-art performance with 85.9\% $mAP_{50}$, 88.2\% $mAP_{50}$, and 79.1\% $mAP_{50}$, respectively. Our code is available at \href{https://github.com/ZiMo-Chen/ProtoHGF}{GitHub}.
\end{abstract}



\maketitle

\section{Introduction}
RGB-Thermal (RGBT) has been widely used in UAV inspection~\cite{chen2025cross}, nighttime traffic surveillance~\cite{pi2020convolutional}, and emergency response~\cite{HOH-Net} by exploiting the complementary strengths of visible textures and thermal cues.
Visible (RGB) images provide rich texture and structural details, but degrade severely under low illumination and occlusion. In contrast, thermal infrared (Thermal) images capture heat-radiation cues and are more robust under adverse conditions, yet often lack clear boundaries and fine-grained appearance details due to limited spatial resolution~\cite{he2023multispectral}. These complementary properties make RGBT detection a promising solution for robust perception in complex environments.

\begin{figure}[htbp]
\centering
\includegraphics[width=0.47\textwidth,height=5.3cm]{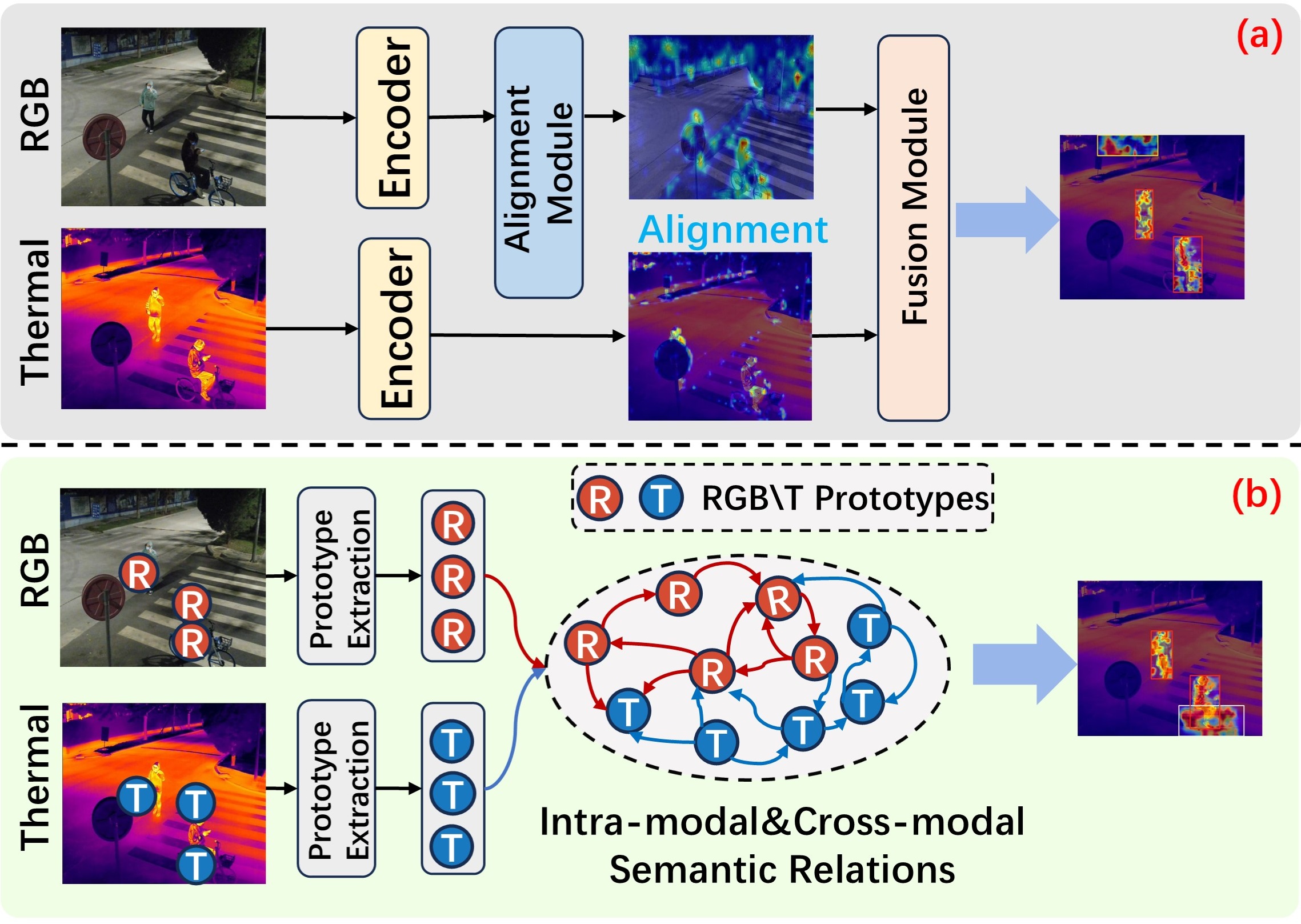}
\caption{
Comparison of cross-modal fusion paradigms for RGBT object detection. (a) Dense full-resolution interaction. (b) ProtoHG-Fusion aggregates target-relevant information into semantic prototypes.
}
\label{motivation-hygraph}
\end{figure}

Most existing RGBT object detection methods are based on cross-modal feature fusion, aiming to fully exploit the complementary information between visible and thermal infrared modalities. To achieve this, a large body of research has focused on the design of complex feature-level fusion architectures~\cite{mgff,mcif,uavd,shen2024icafusion} and pixel-level fusion strategies~\cite{Liu_2025_CVPR,bai2024task,zhang2023superyolo}.
However, these dominant paradigm still relies on dense interaction over full-resolution feature maps. 
In other words, it implicitly \textit{assumes that all spatial locations and all modal responses should participate in cross-modal fusion.}
This assumption may subject the model to two key challenges.

\textbf{First}, dense cross-modal interaction over full-resolution feature maps is prone to introducing substantial background interference. In real-world scenes, target regions usually occupy only a small portion of the spatial positions, while a large number of feature responses come from the irrelevant background. As illustrated in \figureautorefname~\ref{motivation-hygraph}(a), when existing methods indiscriminately interact across the entire feature map, background regions also participate in information propagation and fusion. This leads to unnecessary background coupling and further causes unstable cross-modal interference. As a result, the target-related features are weakened, such that the model is limited to fusing discriminative semantics. In other words, existing dense interaction mechanisms favor broad fusion but lack selective aggregation of target-relevant information.

\begin{figure}[htbp]
\centering
\includegraphics[width=0.45\textwidth,height=5.4cm]{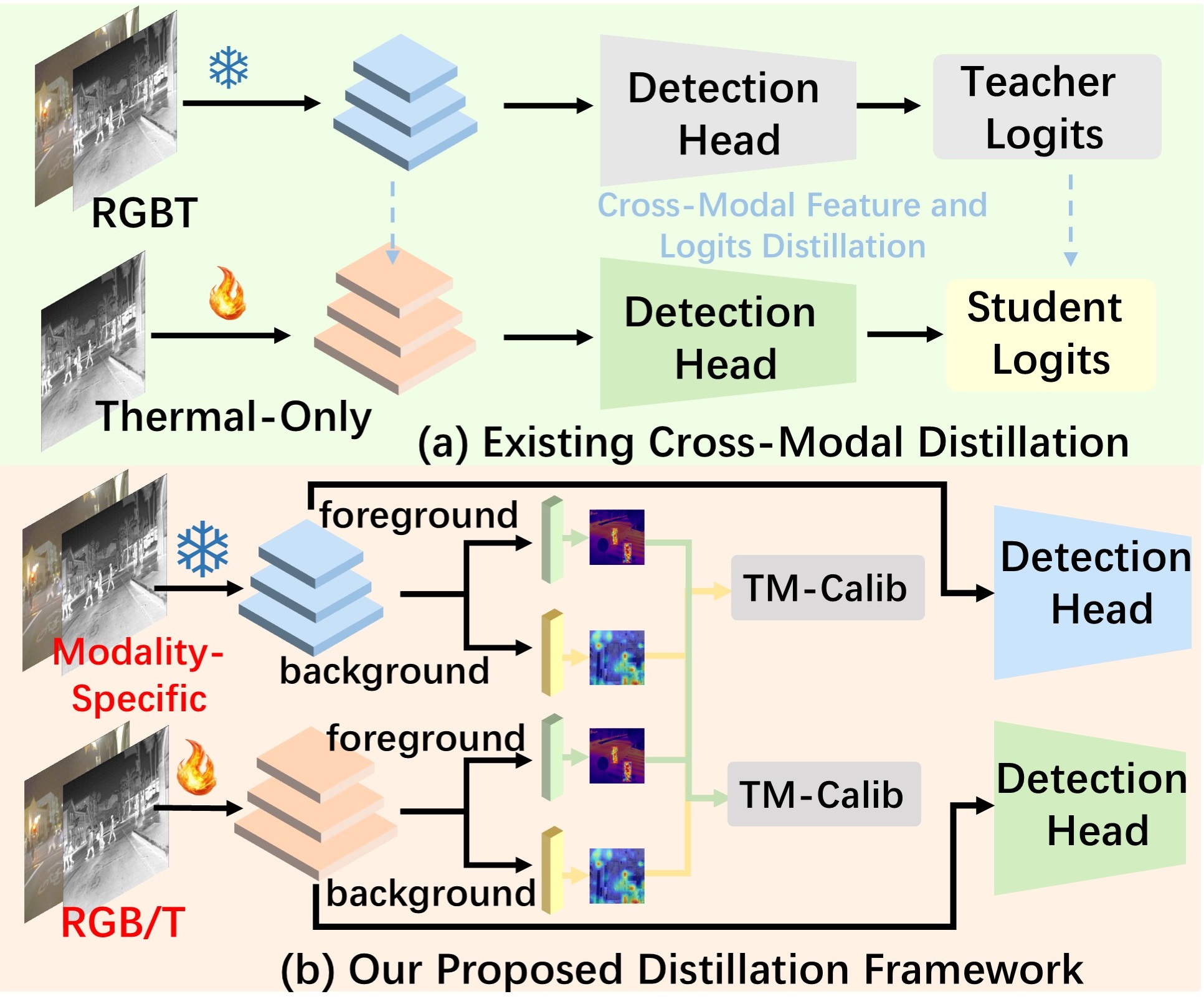}
\caption{
Comparison of distillation paradigms for RGBT detection.
(a) Distillation from a fused multimodal teacher.
(b) TM-Calib uses modality-specific teachers and target-aware masks for foreground-focused pre-fusion calibration.
}
\label{motivation-distill}
\end{figure}

\textbf{Second}, existing distillation paradigms lack effective target-oriented calibration before fusion, making it difficult to provide clean and foreground-focused modal representations for cross-modal fusion. Some methods~\cite{CMDistill,AMFD} attempt to enhance dual-modal feature learning through teacher-student distillation. However, the common practice usually relies on a multimodal teacher after fusion for supervision. As shown in \figureautorefname~\ref{motivation-distill}(a), when the teacher features themselves still contain strong background responses, such background-dominated information may also be transferred to the student network during distillation. 
This issue becomes more severe in scenes with complex backgrounds, weak targets, or imbalanced modality quality. 
Therefore, another important challenge in RGBT detection is how to obtain more target-focused and background-suppressed modal features before fusion.

To address the above limitations, we propose the \textbf{Proto}type \textbf{H}yper\textbf{G}raph \textbf{F}usion \textbf{Net}work (ProtoHGF-Net), a novel framework that redefines cross-modal fusion as prototype-level semantic interaction. 

This framework contains two complementary components: \textbf{Proto}type \textbf{H}yper\textbf{G}raph \textbf{Fusion} (ProtoHG-Fusion) and \textbf{T}eacher-\textbf{M}ask \textbf{Calib}ration Distillation (TM-Calib). 
ProtoHG-Fusion serves as the core fusion module, while TM-Calib serves as a pre-fusion calibration module.
Specifically, ProtoHG-Fusion addresses the first limitation by moving cross-modal interaction from dense spatial maps to a compact prototype-level semantic space. 
It compresses modality-specific features into a small set of semantic prototypes and models their intra-modal structures and cross-modal relations via sparse hypergraph propagation, as illustrated in \figureautorefname~\ref{motivation-hygraph}(b). 
In this way, information exchange is carried out among target-relevant semantic units rather than all spatial positions, which reduces unnecessary background coupling during fusion.
TM-Calib addresses the second limitation by improving the quality of modality features before prototype construction. 
Different from fused-teacher distillation, TM-Calib employs frozen modality-specific teachers to generate target-aware masks and performs foreground-focused intra-modal calibration for RGB and Thermal branches independently, as shown in \figureautorefname~\ref{motivation-distill}(b). 
By suppressing background-dominant responses before fusion, TM-Calib provides more target-focused features for prototype construction and improves the effectiveness of subsequent prototype-level interaction.

Overall, ProtoHGF-Net follows a simple principle: \emph{first calibrate modality features to suppress irrelevant background responses, and then perform cross-modal fusion in a compact prototype-level semantic space}. 
By jointly addressing the granularity of cross-modal interaction and the quality of pre-fusion modality features, ProtoHGF-Net reduces background interference and enables more effective exploitation of RGB-Thermal complementarity.


\section{Related Works}
\subsection{RGBT Object Detection}
Existing RGBT object detection methods can be broadly categorized into two lines. 
The first line focuses on improving multimodal fusion to better exploit complementary information from RGB and Thermal modalities. 
Representative methods mainly enhance cross-modal feature interaction through tailored fusion mechanisms. 
Chen et al.~\cite{mgff} propose MGFF, which uses mask guidance to fuse frequency-domain features. 
Jin et al.~\cite{cssfdet} present CSSFDet, which employs a state-space mechanism to dynamically adjust contextual regional features for fusion.
The second line aims to alleviate weak alignment in RGBT object detection. 
For instance, Zhao et al.~\cite{RGFNet} propose RGFNet, which exploits illumination-invariant properties to align and calibrate cross-modal features. 
Gao et al.~\cite{Uavmatch} adopt multi-scale features together with a Transformer-based alignment network to achieve spatial alignment between modalities. 
OAFA~\cite{oafa} projects features into a common subspace and learns offsets via deformable convolution to perform feature alignment.
Although these methods have achieved promising performance, many of them still rely on dense spatial interaction over full-resolution feature maps. 
Such dense fusion may unnecessarily involve background responses and introduce unstable cross-modal interference, especially in cluttered scenes or under weak correspondence. 
In contrast, our method moves multimodal interaction from full-resolution maps to a prototype-level semantic space, enabling more target-aware and selective cross-modal collaboration.

\subsection{Cross-Modal Knowledge Distillation}
Cross-modal knowledge distillation aims to transfer knowledge learned from one modality to another (the student modality), and it has been widely adopted across various vision tasks. For example, Ma et al.~\cite{ma2024learning} formalize the misalignment of knowledge between modalities and proposed a meta-matching strategy to reduce modality gaps before transfer. In early object detection studies, distillation was typically achieved by learning RoI features~\cite{li2017mimicking} from a teacher model, or by mimicking the teacher’s intermediate representations and soft logits~\cite{chen2017learning}. More recently, researchers have leveraged large-scale models to distill rich prior knowledge into lightweight detectors. Li et al. propose SemFusion~\cite{li2025sam}, which exploits the abundant semantic priors of SAM~\cite{kirillov2023segment} to facilitate knowledge acquisition for object detectors.  Zhao et al.~\cite{m2d-lif} design unimodal distillation together with an illumination-aware module to enhance unimodal feature learning. C2KD~\cite{huo2024c2kd} further adopts a dynamic selection mechanism to mine cross-modal information from non-target categories. Nevertheless, distilling dense representations may also propagate background-dominant responses together with target cues, making it less effective for learning cleaner object-centric semantics prior to cross-modal fusion.

\subsection{Hypergraph Learning Methods}
Hypergraphs~\cite{gao2022hgnn+} connect multiple vertices simultaneously via hyperedges, naturally capturing group interactions and higher-order relations. Recent studies have focused on propagation mechanisms and normalization designs for deep hypergraph models. Dong et al.~\cite{dong2020hnhn} propose HNHN, which treats hyperedges as learnable neurons and improves efficiency and generalization through an adjustable normalization scheme. Feng et al.~\cite{feng2019hypergraph} introduce HGNN, which performs hyperedge-based convolution to propagate and aggregate features across hypergraph structures while preserving the ability to model high-order correlations. 
Subsequent work further incorporates attention mechanisms and set-function perspectives to enhance expressiveness and robustness. 
More recently, Han et al.~\cite{han2023vision} present Vision HGNN, applying HGNN to visual tasks by treating image patches as nodes and explicitly modeling high-level interactions within backbone networks. Later, Feng et al.~\cite{feng2024hyper} integrate hypergraphs into the YOLO architecture, using hypergraph propagation to capture more complex high-order feature relations. Different from prior hypergraph methods that primarily focus on generic visual representation learning, our work explores hypergraph-based relation modeling in RGBT object detection, where hypergraphs are used to support prototype-level cross-modal interaction rather than dense full-resolution fusion.

\section{Method}
\subsection{Overview}
\begin{figure*}[htbp]
\centering
\includegraphics[width=0.98\textwidth,height=9cm]{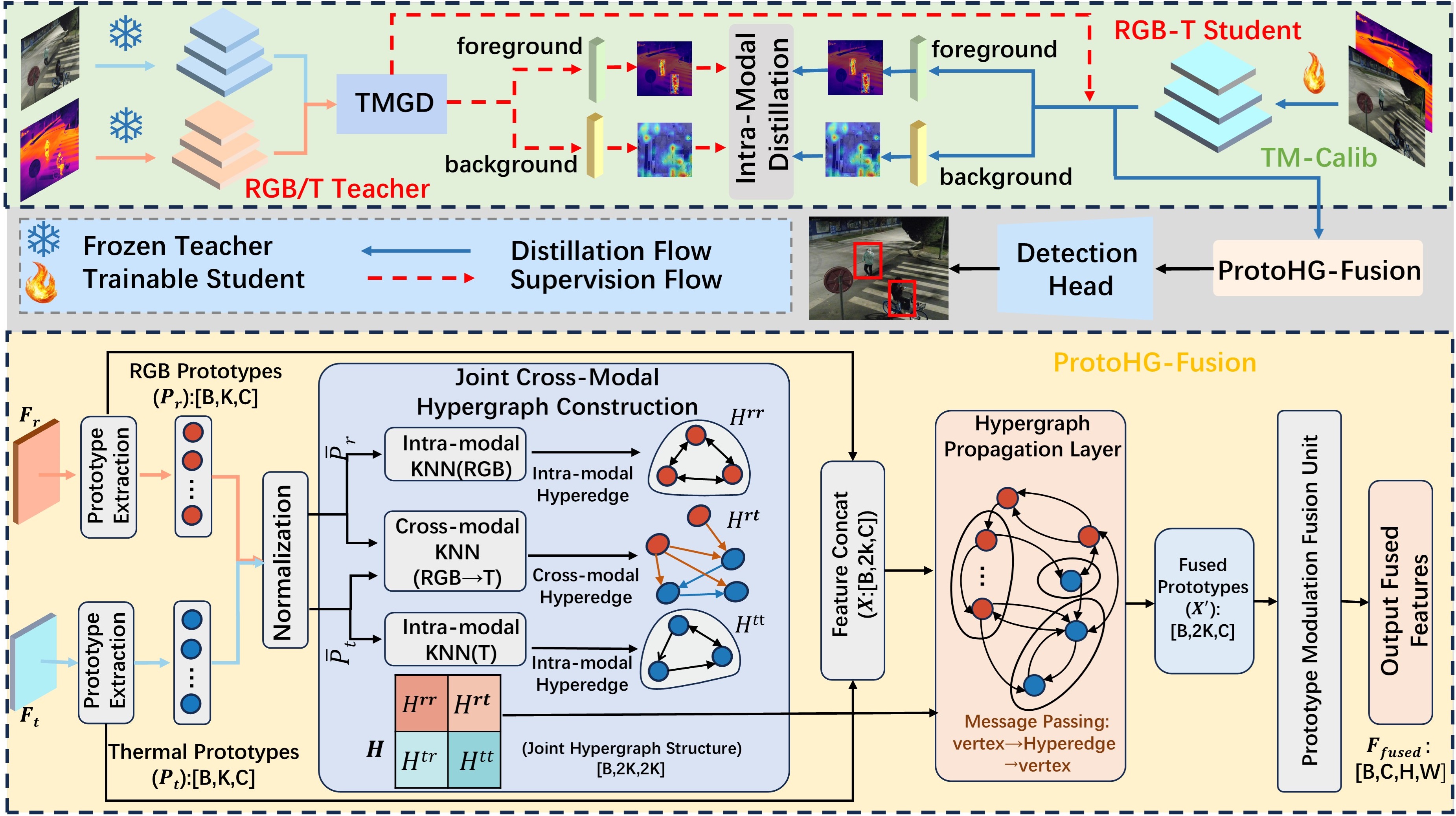}
\caption{
Overview of the ProtoHGF-Net framework, which integrates a ProtoHG-Fusion,  and TM-Calib for RGBT  object detection. Among, TMGD refers to the Teacher-Mask Guided Decomposition operation.
}
\label{framework}
\end{figure*}

As shown in Figure~\ref{framework}, we propose \textbf{ProtoHGF-Net} for RGBT object detection. 
ProtoHGF-Net is built upon a dual-branch backbone and contains two key designs: \textbf{ProtoHG-Fusion}, which conducts prototype-level cross-modal interaction for multimodal fusion, and \textbf{TM-Calib}, which performs target-aware pre-fusion calibration for each modality. 
Given paired RGB and Thermal images, the backbone extracts multi-scale features. 
ProtoHG-Fusion is used to transform these features into compact semantic prototypes and model their interactions via sparse hypergraph propagation, producing fused representations for the detection head. To improve the quality of features before fusion, TM-Calib employs frozen unimodal teachers to generate target-aware masks and calibrate modality-specific representations by emphasizing foreground regions and suppressing background responses.
In our implementation, we adopt a dual-branch RGBT detector based on YOLOv8 as the student, and employ two frozen unimodal detectors, namely an RGB-only detector and a Thermal-only detector, as teachers.

\subsection{Prototype HyperGraph Fusion}
Given the RGBT feature maps extracted by the backbone network, $\mathbf{F}_{\mathrm{r}}, \mathbf{F}_{\mathrm{t}}$ $\in$ $\mathbb{R}^{C \times H \times W}$, C, H, and W represent the number of channels and width/height of the feature map. We propose the ProtoHG-Fusion module to enable stable cross-modal complementary fusion. The key idea of ProtoHG-Fusion is to compress pixel-level features into a small set of semantic prototypes, model intra-modal structures and cross-modal correspondences only at the prototype level, and then feed the prototype-level interactions back to the full feature maps in a modulation manner. Finally, a lightweight gating mechanism is applied to produce the fused output. 
\subsubsection{Prototype Extraction}
We first extract $K$ semantic prototypes from the features of each modality (default $K$ is 6). Specifically, we employ a simple convolutional block composed of convolution and normalization layers to generate $K$ attention maps, and apply a softmax over the spatial dimension to obtain weights $a^k_m$ $\in$ $\mathbb{R}^{H \times W}$ ($m$ $\in$ \{${r}$, ${t}$\}). The $k$-th prototype vector is defined as
\begin{align}
\label{eq1}
    p^k_m=\sum_{u=1}^{H}\sum_{v=1}^{W} a^k_m(u,v)\cdot F_{m}(:,u,v),\quad m\in\{{r},{t}\},
\end{align}
where $F_{m}(:,u,v)\in\mathbb{R}^{C}$ denotes the channel feature of modality $m$ at spatial location $(u,v)$, and $a^k_m(u,v)$ is the contribution weight of that location to the $k$-th prototype. Stacking all prototypes yields $P_{m}=[P^1_{m},\ldots,P^K_{m}] \in \mathbb{R}^{K \times C}$. This prototype representation compresses the $HW$ spatial positions into $K$ semantic nodes, thereby substantially reducing the scale of subsequent cross-modal relation modeling.

\subsubsection{Hypergraph Construction and Propagation}
We construct a joint relation matrix $\mathbf{H} \in \mathbb{R}^{2K \times 2K}$ over the prototypes to capture both intra-modal semantic structures and cross-modal correspondences. 
Specifically, we concatenate the prototypes from the two modalities as $\mathbf{X}=[\mathbf{P}_{\mathrm{r}}; \mathbf{P}_{\mathrm{t}}]\in \mathbb{R}^{2K\times C}$, where $[\cdot;\cdot]$ denotes concatenation along the prototype dimension. 
The resulting relation matrix is block-structured:
\begin{equation}
\mathbf{H} =
\begin{bmatrix}
\mathbf{H}^{rr} & \mathbf{H}^{rt} \\
\mathbf{H}^{tr} & \mathbf{H}^{tt}
\end{bmatrix},
\end{equation}
where $\mathbf{H}^{rr}, \mathbf{H}^{tt} \in \mathbb{R}^{K\times K}$ represent intra-modal connections for the RGB and Thermal modalities, respectively, and $\mathbf{H}^{rt}, \mathbf{H}^{tr}\in \mathbb{R}^{K\times K}$ represent cross-modal connections.

\noindent\textbf{Intra-modal connections.}
To facilitate each modality's cosine similarity computation, we first apply $\ell_2$ normalization to the prototypes, i.e., $\bar{\mathbf{P}}_{m}=\mathrm{Norm}(\mathbf{P}_{m})$, and construct intra-modal connections via $k$-nearest neighbors (KNN). 
For each prototype in modality $m$, we connect it to the $k_{\mathrm{intra}}$ most similar prototypes within the same modality and retain self-loops to stabilize propagation:
\begin{equation}
\mathbf{H}_{i,j}^{mm} =
\begin{cases}
1, & j \in \mathrm{TopK}(\mathbf{S}^{mm}_{i,:}, k_{\mathrm{intra}})\ \text{or}\ j=i, \\
0, & \text{otherwise},
\end{cases}
\end{equation}
where $m\in\{\mathrm{r},\mathrm{t}\}$ and $i,j \in \{1,\dots,K\}$ denote prototype indices. 
Here, $\mathbf{S}^{mm}=\bar{\mathbf{P}}_m\bar{\mathbf{P}}_m^\top\in\mathbb{R}^{K\times K}$ is the intra-modal cosine similarity matrix, and $\mathrm{TopK}(\cdot,k)$ returns the indices of the top-$k$ values in each row. 
$k_{\mathrm{intra}}$ controls the intra-modal connectivity ( $k_{\mathrm{intra}}=3$). 
We define $\mathbf{H}^{mm}_{i,j}=1$ to indicate message passing from prototype $j$ (source column) to prototype $i$ (target row).

\noindent\textbf{Cross-modal connections.}
Cross-modal relations are constructed from the similarity matrix
$\mathbf{S}^{rt}=\bar{\mathbf{P}}_{r}(\bar{\mathbf{P}}_{t})^{\top}\in\mathbb{R}^{K\times K}$.
For each RGB prototype (row index $i$), we select the $k_{\mathrm{cross}}$ Thermal prototypes (column indices $j$) with the highest similarity scores to form cross-modal connections:
\begin{equation}
\mathbf{H}_{i,j}^{rt} =
\begin{cases}
1, & j \in \mathrm{TopK}(\mathbf{S}^{rt}_{i,:}, k_{\mathrm{cross}}),\\
0, & \text{otherwise},
\end{cases}
\label{eq:cross_knn}
\end{equation}
we further set $\mathbf{H}^{tr}=(\mathbf{H}^{rt})^{\top}$ to obtain bidirectional cross-modal links, where $k_{\mathrm{cross}}$ controls the sparsity of cross-modal interactions (default: $k_{\mathrm{cross}}=3$). 
This block structure allows $\mathbf{H}$ to jointly encode intra-modal semantic neighborhoods and cross-modal correspondence neighborhoods, thereby providing a structural prior for subsequent prototype propagation. 
The sparse top-$k$ connectivity also restricts the propagation scope, alleviating noise accumulation and over-smoothing caused by dense relations.

Based on $\mathbf{H}$, we perform two-step prototype propagation in a ``vertex $\rightarrow$ hyperedge $\rightarrow$ vertex'' manner:
\begin{equation}
    \mathbf{E}=\mathrm{Agg}(\mathbf{X},\mathbf{H}^{\top}),\qquad 
    \mathbf{X}'=\mathrm{Agg}(\mathbf{E},\mathbf{H}),
\end{equation}
where $\mathrm{Agg}(\cdot,\cdot)$ is a normalized mean aggregation operator that averages the source features connected to each target according to the relation matrix and normalizes them by the corresponding degree. 
The resulting $\mathbf{X}' \in \mathbb{R}^{2K\times C}$ represents the propagated prototype features. 
To preserve the original semantics and alleviate over-smoothing, we adopt a residual connection:
\begin{equation}
    \tilde{\mathbf{X}}=\mathbf{X}+\mathbf{X}'.
\end{equation}
We then split $\tilde{\mathbf{X}}$ along the prototype dimension to obtain the updated modality-specific prototypes $\tilde{\mathbf{P}}_{\mathrm{r}}, \tilde{\mathbf{P}}_{\mathrm{t}} \in \mathbb{R}^{K\times C}$.

\subsubsection{Prototype Hypergraph-Guided Fusion}
Since the propagated prototypes encode cross-modal interaction information, we use them to generate modulation parameters for recalibrating the full feature maps. For modality $m$, we flatten the propagated prototypes $\mathbf{\tilde{P}}_{m}$ into $z_{m}\in\mathbb{R}^{KC}$ and apply linear projections to obtain channel-wise modulation parameters:
\begin{equation}
    \mathbf{\gamma}_{m}=f_{\gamma}(z_{m}),~ \beta_{m}=f_{\beta}(z_{m}),~
\mathbf{\hat{F}}_{m}=\mathbf{F}_{m}\odot (1+\gamma_{m})+\beta_{m},
\end{equation}
where, $f_{\gamma}(\cdot)$ and $f_{\beta}(\cdot)$ are learnable linear mappings, $\gamma_{m},\beta_{m}\in\mathbb{R}^{C\times 1\times 1}$, and $\odot$ denotes element-wise multiplication. This modulation allows the prototype-level propagation outcomes to influence global channel responses with low computational cost, thereby enhancing cross-modal complementarity.

Finally, we employ a lightweight global module to fuse the two modalities' features:
\begin{equation}
\begin{aligned}
\relax [w_{\text{r}}, w_{\text{t}}] &= \text{Softmax}\left( \psi \left( [\text{GAP}(\phi(\mathbf{\hat{F}} _{\text{r}})); \text{GAP}(\phi(\mathbf{\hat{F}}_{\text{t}}))] \right) \right), \\
F_{\text{fused}} &= w_{\text{r}} \cdot \mathbf{\hat{F}}_{\text{r}} + w_{\text{t}} \cdot \mathbf{\hat{F}}_{\text{t}}.
\end{aligned}
\end{equation}
In this formulation, $\phi$ denotes a $1\times 1$ channel transformation followed by normalization, $\mathrm{GAP}(\cdot)$ is global average pooling, and $\psi(\cdot$) is a two-layer MLP used to predict the modality weights. The scalars $w^{\mathrm{r}}$ and $w^{\mathrm{t}}$ correspond to the weights of the RGBT modalities, respectively. Compared with pixel-wise gating, this global gating strategy is more stable: it can adaptively select the dominant contribution from RGBT across diverse scenes, while avoiding perturbations in detection ranking induced by local noise.

\subsection{Teacher-Mask Calibration Distillation}
\subsubsection{Teacher-Mask Guided Decomposition}
To tackle background interference and modality misalignment in RGBT object detection, TM-Calib uses modality-specific teacher networks to generate target-aware masks. This suppresses irrelevant background information and ensures each modality learns cleaner, object-centric features, reducing misalignment before fusion.

Specifically, we denote the teacher and student features as $\mathbf{F}_m^{T}$ and $\mathbf{F}_m^{S}$, respectively. TM-Calib first generates target-relevant masks from the teacher features. We compute channel and spatial attention and fuse them, and further introduce a lightweight refinement module for calibration:
\begin{equation}
\begin{aligned}
    \mathbf{A}_c^{m} &= \mathrm{CA}(\mathbf{F}_m^{T}), \quad \quad \ \ \
\mathbf{A}_{sg}^{m} = \mathrm{SG}(\mathbf{F}_m^{T}), \\
\mathbf{A}^{m} &= \mathrm{clip}\left(\mathbf{A}_c^{m} \odot \mathbf{A}_\mathbf{sg}^{m} \odot \rho(\mathbf{F}_m^{T}), 0, 1\right),
\end{aligned}
\end{equation}
where $\mathbf{A}_c^{m}\in[0,1]^{C\times 1\times 1}$ is the channel attention~\cite{hu2018squeeze}, $\mathbf{A}_{sg}^{m}\in[0,1]^{1\times H\times W}$ is the spatial attention (SpatialGate), and $\rho(\cdot)$ denotes a two-layer $1\times 1$ refinement network; $\odot$ indicates element-wise multiplication, and $\mathrm{clip}(\cdot)$ truncates values to [0,1]. The resulting $\mathbf{A}^{m}\in[0,1]^{C\times H\times W}$ is a joint channel–spatial mask used to extract foreground responses, while $\mathbf{A}_{sg}^{m}$ serves as a pure spatial prior that enables more direct modeling of background-region weights.
We then use $\mathbf{A}^{m}$ to decompose both the teacher and the student features into foreground and background components:
\begin{equation}
    \begin{aligned}
\mathbf{F}_{\mathrm{fg}}^{(T,m)} &= \mathbf{A}^{m}\odot \mathbf{F}_m^{T}, &\mathbf{F}_{\mathrm{bg}}^{(T,m)} &= \mathbf{F}_m^{T}-\mathbf{F}_{\mathrm{fg}}^{(T,m)},\\
\mathbf{F}_{\mathrm{fg}}^{(S,m)} &= \mathbf{A}^{m}\odot \mathbf{F}_m^{S},  &\mathbf{F}_{\mathrm{bg}}^{(S,m)} &= \mathbf{F}_m^{S}-\mathbf{F}_{\mathrm{fg}}^{(S,m)},
\end{aligned}
\end{equation}
where $\mathbf{F}_{\mathrm{fg}}^{(T,m)}$ and $\mathbf{F}_{\mathrm{bg}}^{(T,m)}$ denote the foreground and background features, respectively. This explicit decomposition allows distillation to be optimized separately for {representation transfer in object regions} and {suppression of background responses}, which better aligns with the characteristics of RGBT imagery, where backgrounds dominate the scene and supervision for small targets is relatively sparse.

\subsubsection{Foreground Alignment Distillation and Background Suppression Distillation} 

Foreground alignment aims to align the student’s target-relevant representations with the teacher’s cleaner foreground representations, while applying constraints only within object regions to avoid negative transfer from the background. We define the foreground alignment loss as a weighted squared feature error:
\begin{equation}
    \mathcal{L}_{\mathrm{fg}}^{m}=\frac{\sum_{(u,v)} \mathbf{W}^{m}(u,v)\left\|\mathbf{F}_{\mathrm{fg}}^{(S,m)}(:,u,v)-\mathbf{F}_{\mathrm{fg}}^{(T,m)}(:,u,v)\right\|_{2}^{2}}{\sum_{(u,v)} \textbf{W}^{m}(u,v)+\epsilon},
\end{equation}
where $(u,v)$ indexes spatial locations, $\epsilon$ is a numerical stability constant (default is $1\times 10^{-6}$), and $\| \cdot\|_{2}^{2}$ denotes the sum of squared errors over the channel dimension. $\mathbf{W}^{m}\in[0,1]^{C\times H\times W}$ is a region-weighting map that emphasizes object areas where the teacher’s responses are more reliable, thereby reducing the dominance of boundary, occluded, or low-confidence regions in the distillation gradients and making distillation more focused on the objects and key regions.

Foreground alignment alone is insufficient to suppress background activations. In RGBT imagery, the background is often complex and covers a large portion of the scene. This can easily induce false positives. We therefore introduce background suppression, which constrains the student’s background energy at locations that the teacher deems as background:
\begin{equation}
    \mathcal{L}_{bg}^m = \frac{\sum_{u,v} (1 - \mathbf{A}_{sg}^m(u, v)) \cdot \left( \frac{1}{C} \| \mathbf{F}_{bg}^{(s,m)}(:, u, v) \|_2^2 \right)}{\sum_{u,v} (1 - \mathbf{A}_{sg}^m(u, v)) + \epsilon} ,
\end{equation}
where $\frac{1}{C} \left \| \cdot \right \|_{2} ^{2}$ measures the energy of background responses, and $1-\mathbf{A}_{sg}^{m}$ acts as a spatial background weighting map. This term complements $\mathcal{L}_{\mathrm{fg}}^{m}$ by aligning object-region representations with the teacher and suppressing irrelevant background activations, thereby alleviating background-dominated distillation bias and reducing false detections.

\subsubsection{Foreground–Background Decoupling Regularization and Overall Objective}
Although background suppression can reduce the magnitude of background responses, foreground and background representations may still remain highly correlated in the feature space, allowing background patterns to leak into the foreground subspace and thus degrade localization accuracy. To further enhance separability, we introduce {Orthogonality Regularization}:
\begin{equation}
    \mathcal{L}_{\mathrm{orth}}^{m}=\left|\cos\big(\mathrm{vec}(\mathbf{F}_{\mathrm{fg}}^{(S,m)}),\mathrm{vec}(\mathbf{F}_{\mathrm{bg}}^{(S,m)})\big)\right|,
\end{equation}
where $\mathrm{vec}(\cdot)$ denotes the flattening operation, and $\cos(\cdot,\cdot)$ is the cosine similarity.

TM-Calib computes and sums the three loss terms for each modality $m\in\{{r},{t}\}$, yielding the overall distillation objective:
\begin{equation}
\mathcal{L}_{\mathrm{tmcalib}}=\sum_{m\in\{{r},{t}\}}\left(\lambda_{\mathrm{fg}}\mathcal{L}_{\mathrm{fg}}^{m}+\lambda_{\mathrm{bg}}\mathcal{L}_{\mathrm{bg}}^{m}+\lambda_{\mathrm{orth}}\mathcal{L}_{\mathrm{orth}}^{m}\right),
\end{equation}
where $\lambda_{\mathrm{fg}}, \lambda_{\mathrm{bg}}, \lambda_{\mathrm{orth}}$ are weighting coefficients, set to 1.0, 2.0, and 0.07 by default, controlling foreground alignment, background suppression, and decoupling regularization, respectively.

In summary, we utilize $\mathcal{L}_{\mathrm{tmcalib}}$ to optimize our TM-Calib. Therefore, for the entire framework of ProtoHGF-Net, the overall loss function is defined as:
\begin{equation}
   \mathcal{L} = \mathcal{L}_{\mathrm{det}}+ \mathcal{L}_{\mathrm{tmcalib}},
\end{equation}
where $\mathcal{L}_{\mathrm{det}}$ represents the loss of classification and localization of the original YOLOv8 model.

\begin{table}[t]
    \caption{Comparison on the DroneVehicle dataset. All methods employ
    oriented bounding box detection heads. The best and second-best
    results among RGB--T methods are highlighted in bold and underlined,
    respectively.}
    \label{table:dronevehicle_results}

    \centering
    \setlength{\tabcolsep}{1.6pt}
    \renewcommand{\arraystretch}{0.93}

    \resizebox{\columnwidth}{!}{%
    \begin{tabular}{@{}lcccccccc@{}}
        \toprule
        \textbf{Method}
        & \textbf{RGB}
        & \textbf{T}
        & \textbf{Car}
        & \textbf{Freight}
        & \textbf{Truck}
        & \textbf{Bus}
        & \textbf{Van}
        & \textbf{$mAP_{50}$} \\
        \midrule

        \multicolumn{9}{l}{\textit{RGB-only methods}} \\

        S$^{2}$ANet~\cite{s2anet}
        & \Checkmark
        & \XSolidBrush
        & 80.0
        & 54.2
        & 42.2
        & 84.9
        & 43.8
        & 61.0 \\

        RoITrans~\cite{roitrans}
        & \Checkmark
        & \XSolidBrush
        & 61.6
        & 55.1
        & 42.3
        & 85.5
        & 44.8
        & 61.6 \\

        Oriented R-CNN~\cite{xie2021oriented}
        & \Checkmark
        & \XSolidBrush
        & 80.1
        & 53.8
        & 41.6
        & 85.4
        & 43.3
        & 60.8 \\

        YOLOv8m~\cite{yolov8}
        & \Checkmark
        & \XSolidBrush
        & 90.9
        & 59.1
        & 72.7
        & 93.9
        & 60.0
        & 75.3 \\

        \midrule
        \multicolumn{9}{l}{\textit{Thermal-only methods}} \\

        S$^{2}$ANet~\cite{s2anet}
        & \XSolidBrush
        & \Checkmark
        & 89.9
        & 54.5
        & 55.8
        & 88.9
        & 48.4
        & 67.5 \\

        RoITrans~\cite{roitrans}
        & \XSolidBrush
        & \Checkmark
        & 89.6
        & 51.0
        & 53.4
        & 88.9
        & 44.5
        & 65.5 \\

        Oriented R-CNN~\cite{xie2021oriented}
        & \XSolidBrush
        & \Checkmark
        & 89.8
        & 57.4
        & 53.1
        & 89.3
        & 45.4
        & 67.0 \\

        YOLOv8m~\cite{yolov8}
        & \XSolidBrush
        & \Checkmark
        & 98.3
        & 77.4
        & 80.2
        & 97.0
        & 68.1
        & 84.2 \\

        \midrule
        \multicolumn{9}{l}{\textit{RGB--T methods}} \\

        SLBAF~\cite{cheng2023slbaf}
        & \Checkmark
        & \Checkmark
        & 90.2
        & 68.6
        & 72.0
        & 89.9
        & 59.9
        & 76.1 \\

        DMM~\cite{zhou2025dmm}
        & \Checkmark
        & \Checkmark
        & 90.4
        & 63.0
        & 77.8
        & 88.7
        & 66.0
        & 77.2 \\

        SemFusion~\cite{li2025sam}
        & \Checkmark
        & \Checkmark
        & 90.4
        & 68.9
        & 80.0
        & 89.9
        & 68.5
        & 79.6 \\

        OAFA~\cite{oafa}
        & \Checkmark
        & \Checkmark
        & 90.3
        & \underline{73.3}
        & 76.8
        & 90.3
        & 66.0
        & 79.4 \\

        MGFF~\cite{mgff}
        & \Checkmark
        & \Checkmark
        & 90.4
        & 69.6
        & 80.9
        & 89.9
        & 68.0
        & 79.8 \\

        UAVD~\cite{uavd}
        & \Checkmark
        & \Checkmark
        & \textbf{98.6}
        & 69.8
        & \textbf{83.9}
        & \underline{96.9}
        & 66.1
        & \underline{83.0} \\

        IGIANet~\cite{chen2026igianet}
        & \Checkmark
        & \Checkmark
        & 90.5
        & 66.2
        & 78.2
        & 90.3
        & \underline{69.4}
        & 80.9 \\

        \textbf{ProtoHGF-Net}
        & \Checkmark
        & \Checkmark
        & \underline{98.5}
        & \textbf{79.5}
        & \underline{83.5}
        & \textbf{97.4}
        & \textbf{70.7}
        & \textbf{85.9} \\

        \bottomrule
    \end{tabular}%
    }

    \Description{Comparison of RGB-only, thermal-only, and RGB--T
    oriented object detection methods on the DroneVehicle dataset.}
\end{table}

\section{Experiments}
\subsection{Experimental Settings}
All experiments are conducted on an NVIDIA DGX A100 GPU. Our proposed ProtoHGF-Net is a dual-backbone branch detector based on YOLOv8m and implemented using the Ultralytics framework~\cite{wan2025yolov11}. The teacher model employs a pre-trained modality-specific detector based on YOLOv8m.



\subsubsection{Datasets and Training details}
We evaluate our method on three RGBT detection benchmarks: DroneVehicle~\cite{dronevehicle}, DVTOD~\cite{dvtod}, and FLIR~\cite{flir}. Unless otherwise specified, all models are optimized using Stochastic Gradient Descent (SGD)~\cite{sgd} with an initial learning rate of 0.001 and a momentum of 0.937, and the input pairs are resized to 640$\times$640.

For DroneVehicle~\cite{dronevehicle}, we adopt an oriented bounding-box detection head and report mean Average Precision ($mAP$) at IoU 0.5 on the val split. The model is trained for 60 epochs with a batch size of 8. During testing, rotated boxes are evaluated using the same IoU computation as MMRotate~\cite{zhou2022mmrotate}, while horizontal boxes follow the default IoU computation in YOLOv8.
For DVTOD~\cite{dvtod}, we use a horizontal bounding-box (HBB) detection head. Following~\cite{dvtod}, evaluation is conducted with an IoU threshold of 0.5 and a confidence threshold of 0.05, reporting both $AP_{50}$ and $mAP$. The model is trained for 150 epochs with a batch size of 16.
For FLIR~\cite{flir}, we also use a horizontal bounding-box detector and evaluate with the standard $mAP_{50}$ and $mAP$ metrics. The model is trained for 36 epochs with a batch size of 16.



\subsection{Comparison with State-of-the-Art Methods}

\subsubsection{Comparison on DroneVehicle.} As shown in Table~\ref{table:dronevehicle_results}, the multimodal approach significantly outperforms unimodal (RGB or thermal) methods. Specifically, compared with the unimodal teacher model (YOLOv8m) trained on either RGB or thermal data, our meth-
od achieves substantially better results (75.3\% vs. 85.9\%; 84.2\% vs. 85.9\%). These findings demonstrate that our method can effectively leverage the two unimodal teacher models to achieve efficient cross-modal fusion.
Moreover, compared with the SOTA method UAVD, our method achieve an overall gain of 2.9\% $mAP_{50}$. These results show that our approach effectively leverages the advantages of distillation and cross-modal fusion, resulting in superior performance. This comparison highlights the advantage of our fusion design.

\begin{table}[t]
\centering
\small
\caption{Comparison on the DVTOD dataset.  }
\begin{tabular}{c|c|c|c|c|c}
\hline
Method & Type & Person & Car & Bicycle & $mAP_{50}(\%)$ \\
\hline
YOLOv5~\cite{yolov5}   & \multirow{3}{*}{R} & 23.3 & 50.2 & 26.7 & 33.4 \\
YOLOv8~\cite{yolov8}   &                      & 28.4 & 52.7 & 24.7 & 35.3 \\
YOLOv10~\cite{wang2024yolov10}  &                      & 30.4 & 50.8 & 23.8 & 35.0 \\
\hline
YOLOv5~\cite{yolov5}   & \multirow{3}{*}{T}  & 86.2 & 75.9 & 86.2 & 82.8 \\
YOLOv8~\cite{yolov8}   &                      & 86.2 & 75.6 & 88.1 & 83.3 \\
YOLOv10~\cite{wang2024yolov10}  &                      & 85.9 & 74.9 & \underline{85.3} & 82.1 \\
\hline
YOLOv5~\cite{yolov5}    & \multirow{6}{*}{R+T} & 88.8 & 74.3 & 74.6 & 79.2 \\
CMX~\cite{zhang2023cmx}       &                         & 88.9 & 75.9 & 79.6 & 81.6 \\
CFT~\cite{cft}        &                         & 88.9 & 78.0 & 81.3 & 82.7 \\
CMA~\cite{dvtod}   & & \underline{90.3} & \underline{81.6} & {83.1} & \underline{85.0} \\
\textbf{Ours} &                & \textbf{92.1} & \textbf{85.3} & \textbf{87.1} & \textbf{88.2} \\
\hline
\end{tabular}
\label{tab:dvtod}
\end{table}

\begin{table}[htbp]
  \centering
  \small
  \caption{Comparison on the FLIR dataset.  }
  \label{table:flir_aligned}
  \begin{tabular}{l |l| c c} \hline
    {Method} & {Type} & $mAP_{50}(\%)$ &  $mAP(\%)$    \\
    \hline
    YOLOv5~\cite{yolov5} & T        &76.2              & 40.8  \\
    YOLOv5~\cite{yolov5}   & R       & 62.7         & 30.4  \\
    YOLOv8~\cite{yolov8}             & T        & 76.4       & 41.1        \\
    YOLOv8~\cite{yolov8}             & R       & 63.9       & 30.4          \\
    YOLOv10~\cite{wang2024yolov10}             & T        & 74.9       & 40.6        \\
    YOLOv10~\cite{wang2024yolov10}             & R       & 58.4       & 28.3          \\
    \hline
    YOLOv5~\cite{yolov5}    & R+T    & 76.9                & 41.1     \\
    CFT~\cite{cft}       & R+T    & 78.7            & 40.2  \\
    CMX~\cite{zhang2023cmx}      & R+T    &\textbf{82.2}               & 41.4     \\
    LRAF-Net~\cite{LRAF}   & R+T    & 80.5                &\underline{42.8} \\
    \textbf{Ours}             & R+T    & \underline{79.1}  & \textbf{45.1} 
    \\ \hline
  \end{tabular}
\end{table}

\subsubsection{Comparison on DVTOD}As reported in Table~\ref{tab:dvtod}, our proposed ProtoHGF-Net also demonstrates outstanding performance on the DVTOD dataset. Compared with unimodal methods, our method shows a clear advantage. Under the thermal-only setting, ProtoHGF-Net surpasses YOLOv5~\cite{yolov5}, YOLOv8~\cite{yolov8}, and YOLOv10~\cite{wang2024yolov10} by 5.4\%, 4.9\%, and 6.1\% in terms of $mAP_{50}$, respectively. Under the multimodal input setting, we outperform CMX~\cite{zhang2023cmx} by 6.6\% $mAP_{50}$, and achieve higher $mAP_{50}$ than CFT~\cite{cft} and CMA~\cite{dvtod} by 5.5\% and 3.2\%, respectively. These results verify the effectiveness of the proposed approach.

\begin{table}[htbp]
    \centering
    \renewcommand{\arraystretch}{0.9} 
    \small
    \caption{Ablation study of different module components on the DVTOD and DroneVehicle datasets. Among them PHG denotes ProtoHG-Fusion, TMC refers to TM-Calib.}
    \begin{tabular}{cc|cc|cc}
        \hline
        \multicolumn{2}{c|}{Component} &
        \multicolumn{2}{c|}{DVTOD} &
        \multicolumn{2}{c}{DroneVehicle} 
        \\
        \hline  
        PHG & TMC & $mAP_{50}(\%)$ & $mAP(\%)$ & $mAP_{50}(\%)$ & $mAP(\%)$ \\
        \hline  
        \XSolidBrush & \XSolidBrush & 86.4 & 55.4 & 84.5 & 55.4 \\
        \Checkmark & \XSolidBrush &   86.8  &  55.8     & 85.1 & 55.5 \\
        \XSolidBrush & \Checkmark & 87.2 & 55.9 & 85.8 & 56.0  \\
        \Checkmark & \Checkmark & \textbf{88.2} & \textbf{56.1} & \textbf{85.9} & \textbf{56.5}  \\
        \hline
    \end{tabular}
    \label{table:ablation}
\end{table}

\subsubsection{Comparison on FLIR}As shown in Table~\ref{table:flir_aligned}, our proposed method remains highly competitive on the FLIR dataset when compared with a wide range of SOTA approaches.  Although our $mAP_{50}$ is slightly lower than that of LRAF-Net~\cite{LRAF} and CMX, the overall $mAP$ better reflects the comprehensive performance of the detector. In this regard, our method surpasses LRAF-Net by 2.3\% in $mAP$ and exceeds CMX by 3.7\% in $mAP$. In general, the results on these three datasets validate the strong generalizability of our approach.

\begin{table}[t]
\centering
\small
\caption{Ablation study of different loss terms.}
\begin{tabular}{cc|c|c|c}
\hline
$mAP_{50}(\%)$ & $mAP(\%)$ & $\mathcal{L}_{\mathrm{fg}}^{m}$ & $\mathcal{L}_{\mathrm{bg}}^{m}$ & $\mathcal{L}_{\mathrm{orth}}^{m}$ \\
\hline
85.1 & 55.5 &  &  &  \\
84.9 & 55.8 & \Checkmark &  &  \\
84.8 & 55.5 &  & \Checkmark &  \\
85.0 & 55.8 &  &  & \Checkmark \\
85.6 & 56.3 & \Checkmark & \Checkmark &  \\
85.5 & 56.1 & \Checkmark &  & \Checkmark \\
\textbf{85.9} & \textbf{56.5} & \Checkmark & \Checkmark & \Checkmark \\
\hline
\end{tabular}
\label{tab:ablation_loss}
\end{table}

\begin{table}[t]
\centering
\caption{Ablation experiments on different distillation strategies.}
\begin{tabular}{c|cc|cc}
\hline
\multirow{2}{*}{Method} & \multicolumn{2}{c|}{DVTOD} & \multicolumn{2}{c}{DroneVehicle} \\
\cline{2-5}
 & $mAP_{50}(\%)$ & $mAP(\%)$ & $mAP_{50}(\%)$ & $mAP(\%)$ \\
\hline
CWD    & 87.0 & 56.1 & 85.6 & 56.1 \\
\hline
PKD & 87.0 & 55.5 & 84.2 & 54.7 \\
\hline
TM-Calib  & \textbf{88.2} & \textbf{56.1} & \textbf{85.9} & \textbf{56.5} \\
\hline
\end{tabular}
\label{tab:distill_method}
\end{table}

\subsection{Ablation Study}
\subsubsection{Study on the Effectiveness of Individual Components.}
As shown in Table~\ref{table:ablation}, we conduct ablation studies on the proposed components on the DVTOD and DroneVehicle datasets to comprehensively validate the effectiveness of our model. Compared with the baseline, introducing the ProtoHG-Fusion module and the TM-Calib strategy improves $mAP$ by 0.4\% and 0.5\%, respectively; meanwhile, $mAP_{50}$ increases from 86.4\% to 86.8\% and 87.2\%. Similar performance gains are observed on the DroneVehicle dataset. These results indicate that the proposed modules and distillation strategy can effectively exploit complementary information from different modalities.

\subsubsection{Study on the Effectiveness of Different Loss Terms.}
To further investigate the effectiveness of the proposed loss design in TM-Calib, we perform a detailed ablation study on the individual loss terms on the DroneVehicle dataset, as shown in Table~\ref{tab:ablation_loss}. While using any single loss term alone improves upon the baseline, combining them yields larger gains. In particular, jointly applying $\mathcal{L}_{\mathrm{fg}}^{m}$ and $\mathcal{L}_{\mathrm{bg}}^{m}$ boosts $mAP_{50}$ from 85.1\% to 85.6\%, and further adding $\mathcal{L}_{\mathrm{orth}}^{m}$ leads to additional improvements, achieving the best performance of 85.9\%. These results indicate that jointly enhancing foreground features, suppressing background interference, and enforcing feature decoupling are crucial for effective modality calibration.

\subsubsection{Study on the Effectiveness of Distillation Strategies.}
To verify that our distillation design is crucial for RGBT detection, we compare TM-Calib with representative distillation baselines under the same detector, backbone, training schedule, and evaluation protocol.
Specifically, we evaluate (i) CWD~\cite{cwd}, which distills dense feature responses in a channel-wise manner, and (ii) PKD~\cite{cao2022pkd}, a prediction/logit-based distillation that transfers supervisory signals from the teacher’s outputs.
In contrast, TM-Calib performs teacher-mask guided, region-aware distillation to calibrate target-relevant representations while suppressing background-driven negative transfer.
As shown in Table~\ref{tab:distill_method}, TM-Calib achieves the best performance on both DVTOD and DroneVehicle. Compared to PKD, it improves $mAP$ by 0.6\% on DVTOD and 1.8\% on DroneVehicle. When compared with CWD, TM-Calib outperforms it by 1.2\% in $mAP_{50}$ on DVTOD and 0.4\% in $mAP$ on DroneVehicle.
These results demonstrate that region-aware distillation, which minimizes background interference, leads to more effective knowledge transfer and enhanced detection performance.

\begin{figure}[htbp]
\centering
\includegraphics[width=0.45\textwidth,height=7.5cm]{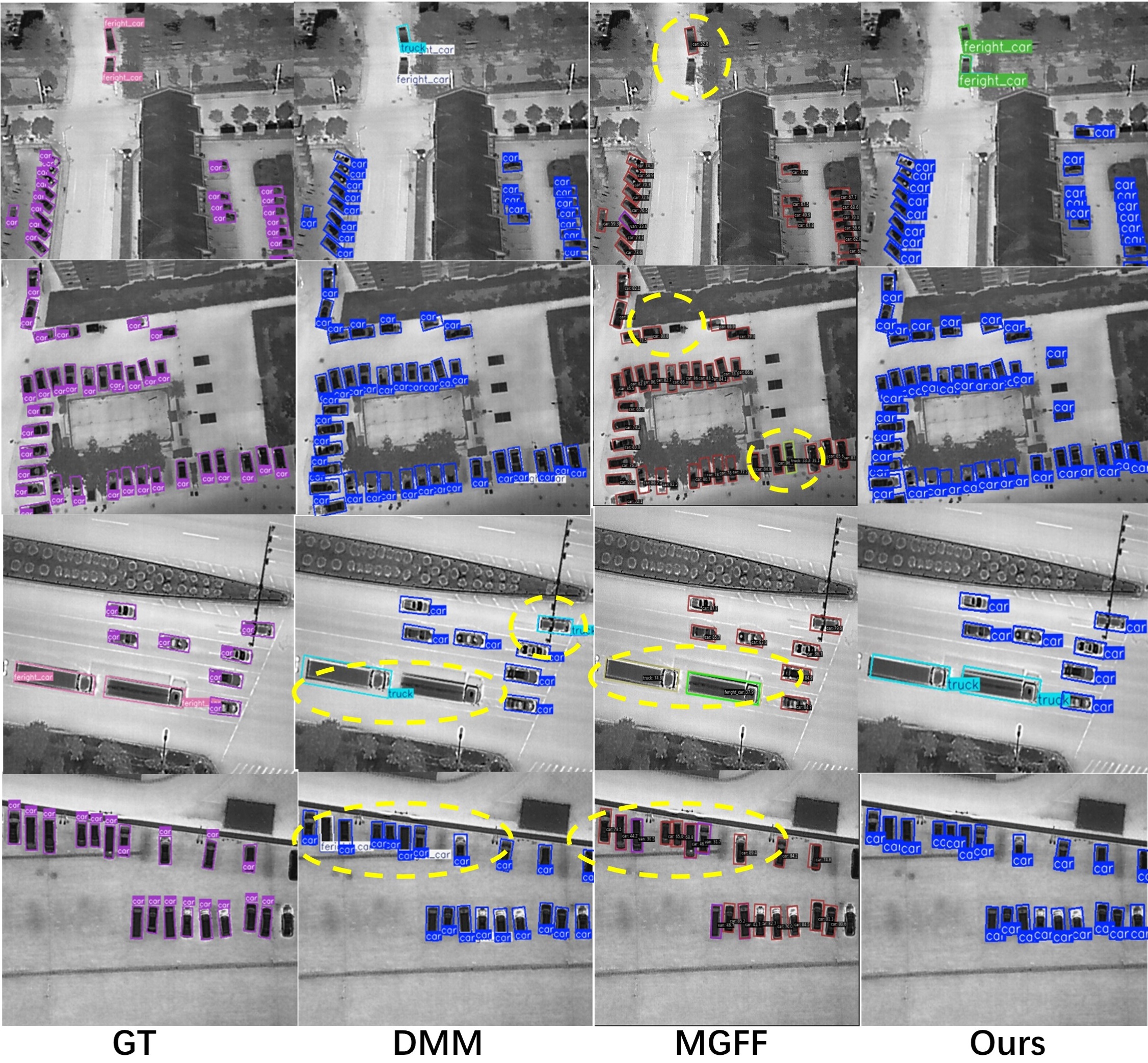}
\caption{Comparison of different detection results, where the yellow dashed line represents detection errors.
}
\label{res_vis}
\end{figure}

\subsubsection{Study on the Effectiveness of the Fusion Method.}
To evaluate the effectiveness of ProtoHG-Fusion, we conduct ablation experiments by replacing it with different fusion strategies. Using Add as the baseline, we compare it with Concat and LIFAdd~\cite{m2d-lif}. As shown in Table~\ref{tab:fusion_method}, ProtoHG-Fusion improves $mAP_{50}$ by 1.8\% on DVTOD and  1.1\% $mAP$ on DroneVehicle over the baseline. 
Compared to the graph-based model, the hypergraph-based formulation consistently performs better. On DVTOD, the proposed hypergraph model improves $mAP_{50}$ by 0.8\%, and similar improvements are observed on DroneVehicle. This highlights that the performance gain stems not only from prototype compression or sparse interactions, but also from hypergraph propagation's ability to capture higher-order dependencies among prototypes.
We further investigate the prototype-level joint relation matrix by comparing the default \emph{hard} top-$k$ construction with a \emph{soft} alternative. Specifically, \textbf{soft intra-modal relations} are computed using masked softmax over the top-$k$ neighbors, while \textbf{soft cross-modal relations} are obtained via softmax on cross-modal connections. As shown in Table~\ref{tab:fusion_method}, the soft-relation variant achieves 86.3\%/$55.7\%$ mAP$_{50}$/mAP on DVTOD and 84.3\%/$55.5\%$ on DroneVehicle, both lower than ProtoHG-Fusion. This suggests that soft relations weaken the sparsity of prototype interactions and introduce less reliable connections. In contrast, the hard 0/1 relation matrix enforces a stronger structural prior and confines message passing to more confident connections, leading to better detection performance. See the appendix for details.
\begin{table}[t]
\centering
\caption{Ablation experiments on different fusion strategies.}
\small
\begin{tabular}{c|cc|cc}
\hline
\multirow{2}{*}{Method} & \multicolumn{2}{c|}{DVTOD} & \multicolumn{2}{c}{DroneVehicle} \\
\cline{2-5}
 & $mAP_{50}(\%)$ & $mAP(\%)$ & $mAP_{50}(\%)$ & $mAP(\%)$ \\
\hline  
Add (baseline)    & 86.4 & 55.4 & 84.5 & 55.4 \\
\hline
Concat & 87.0 & 54.9 & 85.2 & 56.2 \\
\hline
LIFAdd & 86.7 & 55 & 85.3 & 56.4 \\
\hline
Soft-relation  & {86.3} & {55.7} & {84.3} & {55.5} 
\\
\hline
Graph-based  & {87.4} & {55.8} & {85.1} & {55.6} \\
\hline
ProtoHG-Fusion  & \textbf{88.2} & \textbf{56.1} & \textbf{85.9} & \textbf{56.5} \\
\hline
\end{tabular}
\label{tab:fusion_method}
\end{table}

\begin{table}[htbp]
\centering
\caption{Cost comparison between the ProtoHGF-Net method and other SOTA methods.
}
\small
\begin{tabular}{c|c|c|c}
\hline
{Methods} & {FLOPs(G)$\downarrow$} & {Params(M)$\downarrow$} & $mAP_{50}$ (\%) $\uparrow$ \\
\hline
CIAN~\cite{cian} & \textbf{70.0} & - &70.8 \\ 
TSFADet~\cite{TSFAdet} &109.0 & {104.0} &73.9 \\ 
C2Former~\cite{yuan2024c2former} & 89.9 & 100.8 & 74.2 \\
CoDAF~\cite{Codaf} & 224.9 & {67.3} & {78.6} \\
MGFF~\cite{mgff} & 114.0 & {123.0} & {79.8} \\
ProtoHGF-Net & 133.38 & \textbf{50.4} & \textbf{85.9}\\
\hline 
\end{tabular}
\label{table:ablation_cost}
\end{table}

\subsection{Computational Cost Comparison}
We compare the computational cost and model size of ProtoHGF-Net with several state-of-the-art RGBT fusion methods, including CIAN~\cite{cian}, TSFADet~\cite{TSFAdet}, C2Former~\cite{yuan2024c2former}, and CoDAF~\cite{Codaf}. As shown in Table~\ref{table:ablation_cost}, ProtoHGF-Net achieves the highest $mAP$ of 85.9\%, with moderate computational cost (133.38 GFLOPs) and a compact model size (50.4M parameters). 
Compared with MGFF (114 GFLOPs, 123M parameters), ProtoHGF-Net outperforms it by +6.1\% in $mAP_{50}$ while using fewer parameters. It also outperforms CoDAF by +7.3\% in $mAP_{50}$, with more than 50\% reduction in FLOPs. 
Despite CIAN's lightweight design (70G FLOPs), it falls short in performance (70.8\% $mAP$). 
These results demonstrate that ProtoHGF-Net offers a strong trade-off between accuracy and efficiency, making it ideal for resource-constrained RGBT object detection tasks.
\begin{figure}[htbp]
\centering
\includegraphics[width=0.45\textwidth,height=4.2cm]{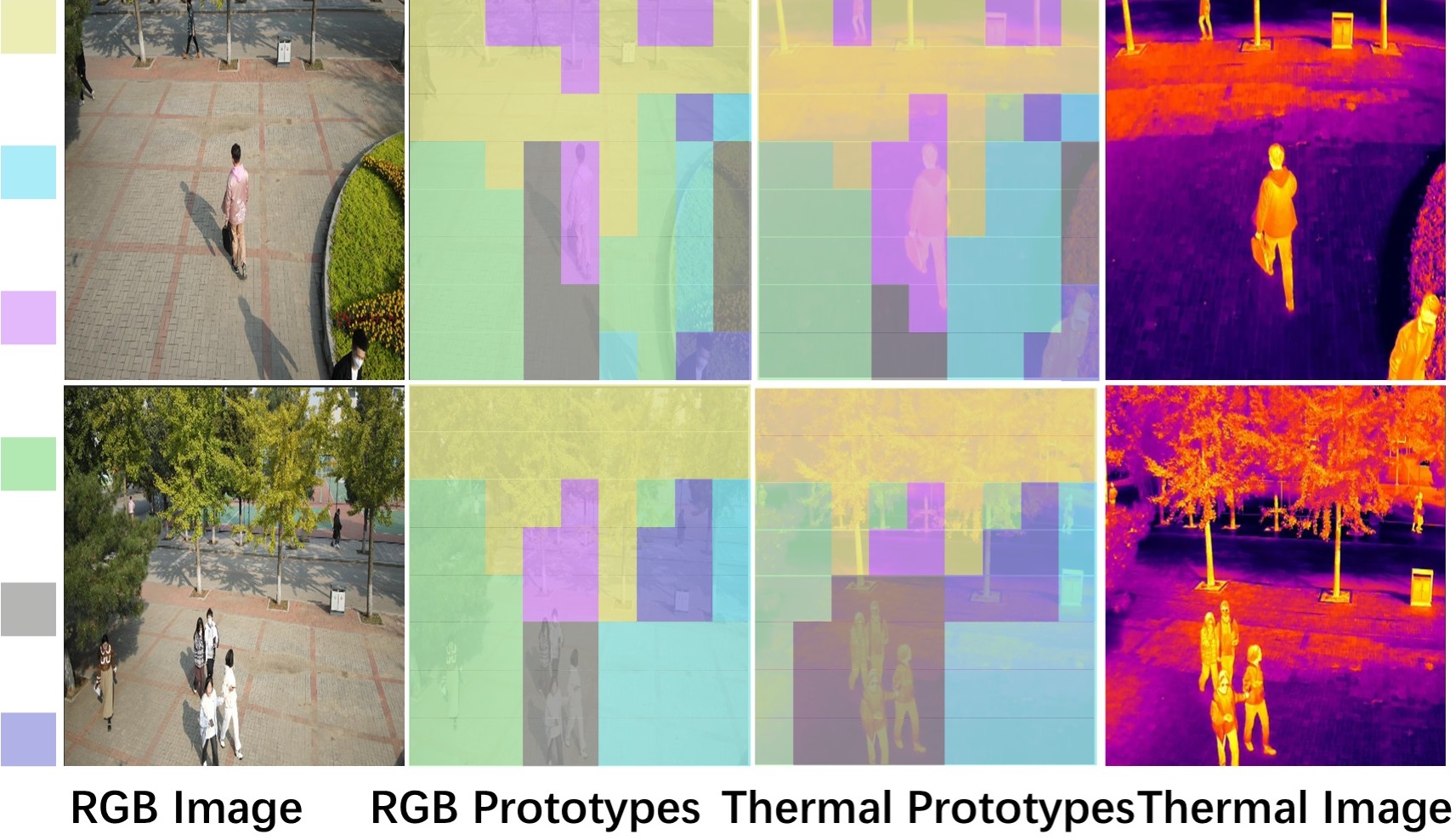}
\caption{Visualization of prototype attention maps. 
}
\label{prototypes}
\end{figure}

\subsection{Visual Analysis}
We provide qualitative comparisons of detection results produced by different methods on the DroneVehicle dataset. As shown in \figureautorefname~\ref{res_vis}, existing approaches are prone to missed detections and false alarms under complex scenes. In contrast, our proposed ProtoHGF-Net effectively enhances foreground target features and yields the fewest false positives. In addition, we visualize the $K$ semantic prototypes in the proposed ProtoHG-Fusion module. As illustrated in \figureautorefname~\ref{prototypes}, each semantic prototype attends to different regions, which facilitates effective cross-modal information interaction.
Therefore, the visualizations demonstrate that ProtoHGF-Net effectively enhances foreground target features and facilitates selective cross-modal interaction, leading to improved detection performance.

\section{Conclusion}
In this paper, we propose ProtoHGF-Net for RGBT object detection, featuring intra-modal calibration distillation and prototype-level structural fusion. ProtoHG-Fusion performs hypergraph propagation over compact prototypes with sparse interactions and gated fusion, which mitigates noise diffusion and mismatched cross-modal message passing. TM-Calib leverages frozen unimodal teachers to generate object masks, enabling foreground-weighted alignment and reducing background-driven negative transfer before fusion. Experiments on multiple RGBT benchmarks verify the effectiveness and generalization of our approach. In future work, we will extend ProtoHGF-Net to more challenging settings, such as multi-spectral or video-based detection with temporal consistency.

\begin{acks}
This work was supported in part by the Project of the China–Mozam-
bique “Belt and Road” Joint Laboratory on Smart Agriculture (No. 2024YFE0214000), in part by the National Natural Science Foundation of China (Nos. 62272419 and 62402449), in part by the Zhejiang Provincial Natural Science Foundation of China (No. LQK26F020003), in part by the Major Program of the Natural Science Foundation of Zhejiang Province (No. LD26F020003), and in part by the Key Project of the Jinhua Science and Technology Bureau (No. 2024-2-015).

\end{acks}

\appendix

\bibliographystyle{ACM-Reference-Format}
\balance
\bibliography{sample-base}

\appendix









\end{document}